# Improving Train Track Safety using Drones, Computer Vision and Machine Learning

*by Kirthi Kumar, Senior at TJHSST and Anuraag Kaashyap, Junior at TJHSST*

# Abstract

Millions of human casualties resulting from train accidents globally are caused by the inefficient, manual track inspections. Government agencies are seriously concerned about the safe operations of the rail industry after series of accidents reported across the USA and around the globe, mainly attributed to track defects. Casualties resulting from track defects result in billions of dollars loss in public/private investments and loss of revenue due to downtime, ultimately resulting in loss of the public's confidence. The manual, mundane, and expensive monitoring of rail track safety can be transformed through the use of drones, computer vision, and machine learning.

The primary goal of this study is to develop multiple algorithms that implement supervised and semi-supervised learning that accurately analyze whether a track is safe or unsafe based on simulated training data of train tracks. This includes being able to develop a Convolutional Neural Network that can identify track defects using supervised learning without having to specify a particular algorithm for detecting those defects, and that the new model would both speed up and improve the quality of the track defect detection process, accompanied with a computer vision image-processing algorithm. Our other goals included designing and building a prototype representation of train tracks to simulate track defects, to precisely and consistently conduct the visual inspection using drones. Ultimately, the goal demonstrates that the state of good repairs in railway tracks can be attained through the use of drones, computer vision and machine learning.

# Project Overview

As we witnessed several metro train derailments in the DC Metro region and the public outcry for safe and reliable public transportation, we decided to investigate and research how technology could be applied to solve this regional problem to support our local community. We reached out to the DC Metro Authority and researched during summer about these train derailments and read several publications from Federal Railroad Administration. During our research, we reviewed the safety and security procedures adopted by the DC Metro, mandated by the regulations, to understand the requirements of track inspection procedures. It became obvious that this problem is not only local to the DC Metro region, but is a global problem where derailments are attributed to manual inspections that are done today once in 2-4 weeks as mandated by the local governments/law enforcement agencies due to budget constraints.

According to the statistics from the Federal Railroad Administration, the US Department of Transportation, and the National Transportation Safety Board, millions of human casualties globally result from track defects causing losses, dissatisfaction of commuters, which ultimately leads to the loss of public's confidence and outcry. This created a dire need for a new automated method of identifying track defects rather than the manual and subjective method that exists today, which inspired our project. We used drones to capture the train track footage and developed a unique algorithm (supervised learning) with computer vision tools (MATLAB) such as Pre-Processing, Extraction, Segmentation, Detection and Post-Processing Noise. This resulted in predicting faults in the simulated train track test cases from trained datasets produced from drone footage along with well-documented inspection results. After running a Likert scale and

# Improving Train Track Safety using Drones, Computer Vision and Machine Learning

variation analysis on the data collected from fifteen test cases, five trials for each test case, our algorithm produced a success rate of 96.09% in detecting track defects.

We further extended the project to study the implementation of VGG16 and a Custom Model using Keras with a TensorFlow backend for processing images and adopting multi-layered Convolutional Neural Networks (CNNs). The existing VGG16 model was adapted using transfer learning specific to track images captured during the project. A custom model was also developed to predict the track safety. We used a CNN model to test the images captured by a drone on model tracks. CNNs are a class of feed-forward neural networks for analyzing images and are inspired by the biological processes occurring within brain's visual cortex. CNNs use minimal pre-processing in comparison with other image classification algorithms. The major advantage of this type of neural network is that we do not need to specify an algorithm to detect the classification. Instead, it just learns from examples.

The drone footage captured were classified as safe or defective tracks by image preparation, building and training the model, and eventually testing the model (using industry standard processes for artificial intelligence/machine learning). These tests were accomplished using Python programming language with open-source common libraries of machine learning. The results from the VGG16 model achieved 62 - 97% training accuracy and 69 - 73% validation accuracy. The custom model, developed as part of this project, showed an incredible promise to identify the track defects achieved 79 - 98% training accuracy and 77 - 89% validation accuracy.

This project can set the industrial revolution for the transportation industry and influence the regulators to revisit the mundane "boots on the ground" train track inspections that are done once in 2-4 weeks, due to budget constraints, to daily monitoring of track defects using machine learning and intelligent machines (drones and VR) without compromising the revenue on maintenance schedules. This is an accurate, unique, and real-time solution that is inexpensive and objective, which aims to proactively problem-solve the growing issue of track safety. Our research promotes a future of using drones to perform track inspections powered by machine learning and image processing. This can ensure safety and security across transits on a daily basis rather than the manual inefficient inspection done once in two to four weeks due to budget and revenue constraints. This could save billions of dollars loss in public/private investments, build better public confidence, help save the environment with less pollution, improve revenue, and most importantly prevent millions of human casualties globally.

This research can be further extended to any parts of the safety and security industry by comparing the state of good with respect to the current state of any site/environment using machine learning, computer vision and augmented reality.

# Background Research

Millions of human casualties resulting from train accidents globally are caused by the inefficient inspection of the aging infrastructure of public transportation done manually by railway safety engineers every two to four weeks. Primarily, the National Transportation Safety Board (NTSB) is seriously concerned about the safe operations of the current rail transit industry after series of accidents reported across the USA, which they attributed to track defects (NTSB, 2016). Casualties resulting from track defects results in a loss of billions of dollars in public and private investments and loss of revenue in the industry due to downtime, ultimately results to the dissatisfaction of commuters and loss of the public's confidence in their local metro or subway, leading to more vehicles on the highway and increased pollution (TSB of Canada, 2014). Thus, the reliable detection of defects in railroad tracks is of great importance for both freight and transit rail safety. The manual, mundane, and expensive monitoring of rail track safety can be greatly improved through recent advancements in technology including drones, image processing systems, computer vision, and artificial intelligence (AI). This led to the research question: is it possible to develop a unique system that accurately identifies hazardous areas through deep learning by implementing computer vision, machine learning (ML), and how accurate is this method of identification?

Currently, there is nothing being used to monitor track safety with drones. The current method is "boots on the ground every week/month" and manual track inspection using rail trucks (Camargo et al., 2011). However, there is technology called the Ground Penetration Radar (GPR) that makes laser guided measurements of track gauge and monitor health of rail track ties such as wood, concrete or metal blocks including density of gravel (Saarenketo, 1994). Although there are companies implementing some support of GPR and laser guided support to monitor the health of rail track infrastructure, they are too expensive, and most times not done due to limited funding. Research has been done in this area before with respect to artificial intelligence in the industry, including methods for detecting faults in train tracks. Gibert et al. (n.d.) did research about a deep multi-tasking inspection robot, which used hardware to detect rail gauge and other faults with the tracks. Nageshwaran et al. (2014) did research on an autonomous vehicle that would perform inspection on the tracks and used a remote station to communicate with it to detect the faults. However, the fault with any existing innovations for inspection is that it requires a vehicle to be on the ground so the trains cannot run at the same time.

Improving Train Track Safety using Drones, Computer Vision and Machine Learning# Summary of Project Implementation

Identifying the real-time condition of rail track support systems with respect to the state of good repairs using low cost drone surveillance will proactively identify faults in railway infrastructures rather than reactively problem-solving through labor-intensive and time-consuming manual visual observations, inspections and record keeping, which would limit accidents, saving lives, investments, and the public image of the industry. We requested Washington Metropolitan Area Transit Authority (WMATA) to provide their research facility to study the rules and regulations of track inspections. During our research, we reviewed and followed through the critical elements of track inspections procedures. We built the simulated tracks based on the scaled down version of these actual tracks with key components that would be inspected during our projects – wooden blocks representing the ties, screws/washers representing the fasteners, and connectors representing the rail extension anchors.

The project started off by extracting frames from videos of simulated tracks taken from the Bebop Parrot 2 drone, preprocess them and highlight "problem areas" with safety issues using the computer vision tools in the MATLAB Software so that railway safety engineers can focus on resolving issues rather than finding them. Additionally, the program built in the project can document all safety issues without bias while trains are running (helping no loss in revenue during monitoring), which will improve record keeping of rail track conditions and hence supplement the trained datasets required to build better unsupervised autonomous systems, providing better security to the public railway transportation industry in a consistent and cost-effective way.

We researched further to use machine learning principles and came across using a convolutional neural network (CNN) model to test the images captured by a drone on model tracks. CNNs are a class of feed-forward neural networks for analyzing images and are motivated by the biological processes occurring within brain's visual cortex. CNNs use minimal pre-processing in comparison with other image classification algorithms (like MATLAB model we developed which posed tedious image preprocessing steps). The major advantage of this type of neural network is that we do not need to specify an algorithm to detect the classification. Instead, it just learns from examples. We started off with the VGG16 based model, an existing model used to classify images into a thousand objects that won the ImageNet Large Scale Visual Recognition Challenge in 2014. This concept of adapting an existing model and modifying it to classify different objects is called transfer learning. This research showed promising results and posed challenges after reviewing the testing results with unacceptable execution times. This was taking too much time to analyze the track defects and results were not conclusive. In particular, one trial began showing signs of overfitting, in which the model memorizes the input data instead of learning the prominent features that distinguish the different categories of images. We learned that one way to reduce overfitting was by introducing dropout, in which a fraction of the model weights (that determine classification) would be deleted every epoch, which seems counterproductive, but reduced reliance on irrelevant factors (Srivastava et al.). Hence, we developed our own custom model that gave a boosted performance in analyzing track defects. Our method is based on the following industry standard processes to implement AI/ML neural networks such as capture the images, build the model, train the model and test the model. We repeated the process steps for various scenario by tweaking different parameters and redesigning the model to improve the overall accuracy and feasibility of these models. We used Keras and TensorFlow for processing images

# Improving Train Track Safety using Drones, Computer Vision and Machine Learning

while implementing CNN. Keras is an open-source neural network library, and TensorFlow is Google's open source library for large-scale machine learning and computation.

The technology created in this project will be extremely useful in the industry given that it can inspect while trains are running and provide an unbiased, real-time safety review of millions of miles of track. Additionally, it can be further augmented to include more hardware, such as the GPR, to make invaluable measurements in a short amount of time. While the expectation of the project is to demonstrate the state of good repairs in railway tracks can be attained through computer vision and machine learning (supervised learning) that can support continuous monitoring and analyzing of computer images captured by drone with trained datasets, it can be further extended to semi-supervised and unsupervised machine learning to ensure highest level of safety and security in the transportation industry that could save millions of lives, improve commuter satisfaction and their confidence in public transportation, and avoid public/private investments loss. This project's further research can bring the public/private industry together to build better autonomous rail track safety systems to guard billions of miles of rail tracks around the globe.

# Design and Implementation using Computer Vision

Since there are various aspects of track defect issues and limited time available for this project, we took the following 15 test cases to simulate and test our results for false positive and false negative conditions (shown in Table 1). The following steps were followed to conduct the experiment so that results can be gathered and analyzed further for test results:

## 1. Project Planning

The project team started this project by reviewing major accidents around the US and the world and investigated the root cause of these accidents. We requested our director of the robotics lab to mentor the project and started reviewing how trains operate. We visited the research laboratory (WMATA) to review causes of these accidents and found out majority of them were result of lack of inspections and maintenance of railway infrastructure, especially train tracks. We did extensive review on rail track inspection and maintenance procedures. We reviewed the inspection and maintenance manual for the metro and understood some of the FTA and NTSB guidelines in rail track inspection procedures and mandated frequencies. We were astonished by how manually intensive these operations were and started thinking about how we can make these inspection procedures automated for better results. We started researching about what is happening in universities and the industry about improving the inspection procedures and provide safety and security for common public so that derailments/accidents can be avoided. In addition, we witnessed some accidents around the USA and around the world watching news and videos while researching these topics and surprised about public outrage, loss of lives/injuries and massive public/private investments. Then, we purchased materials including a drone and MATLAB Software with computer vision and image processing toolset so that we can test our hypothesis.

## 2. Building the Train Track Simulation Model

We started with a 3D modeling of rail track using a 3D printer from the lab and then created simulated rail tracks using wooden blocks to represent the wooden ties and cast-iron metal to represent the rail. We soldered the rail to the rail anchors and then drilled holes in these anchors to hold the wooden blocks. The wooden blocks were then riveted to these rail anchors using screws and washers (representing fasteners). Finally, the connectors were prepared to represent rail extension anchors and were placed on these tracks at the end points. We prepared 4 such rail tracks. One was considered as standard track (state of good repairs, control) that was used to compare with another track with simulated defects (problem track, variable). The other two tracks were connected, and a rail truck was placed on it (prepared for the project using wheels, spring, bearings and brackets to simulate real-life rail system). The blue tarp was placed on the floor and gravel were laid to simulate rail track system. The Figures 1 and 2 illustrate the train tracks along with the flying drone to demonstrate the intricacies of the project.

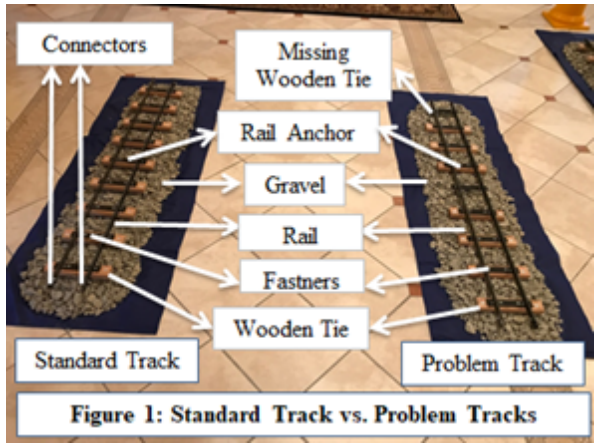 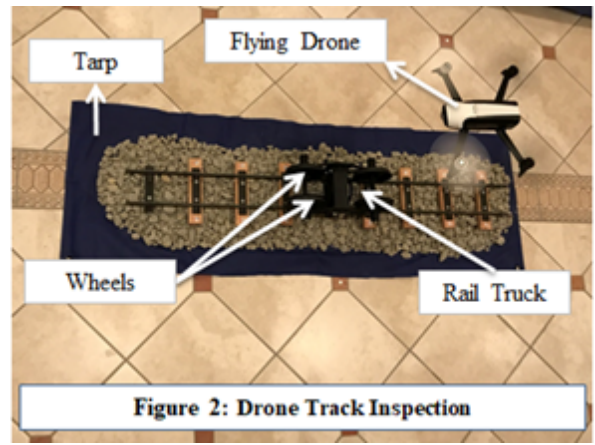

Figure 1: Standard Track vs. Problem Tracks

Figure 2: Drone Track Inspection

## 3. Visual Inspection Using Drone

The following procedures were followed to perform the visual inspection of rail track system with the standard track with no defects: 3.1) The simulated track was placed on the gravel; 3.2) The drone was made to fly over rail track system five times with prescribed coordinates (starting point, height, length and ending point); 3.3) Video and still images were taken from drones were downloaded with proper labels. The same steps were followed, and drone was run over 5 times for all problem track cases. These cases were labelled from 02 – 15 and tabulated in table 1 and 2 listed in the results section.

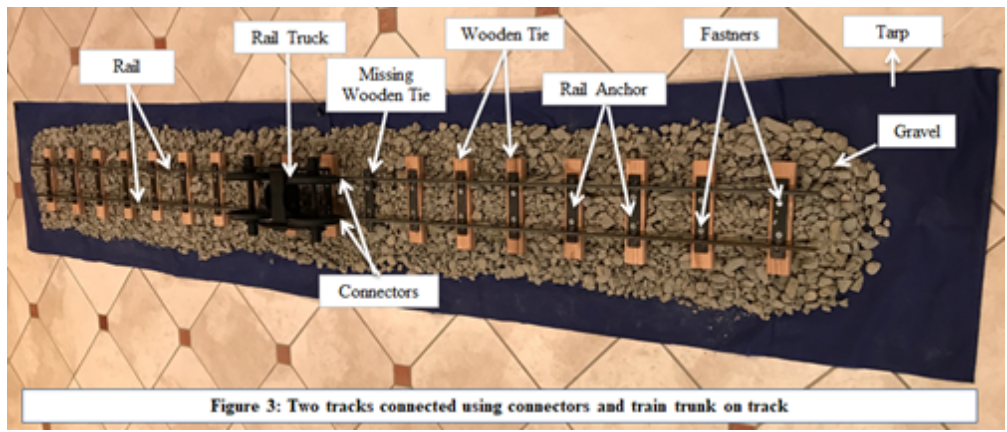

Figure 3: Two tracks connected using connectors and train trunk on track

The track geometry was labeled so that computer vision algorithms can analyze the problem track image and compare with the standard track image for false positive and false negative results. Figure 4 illustrates the labeled track components considered for defect detection.

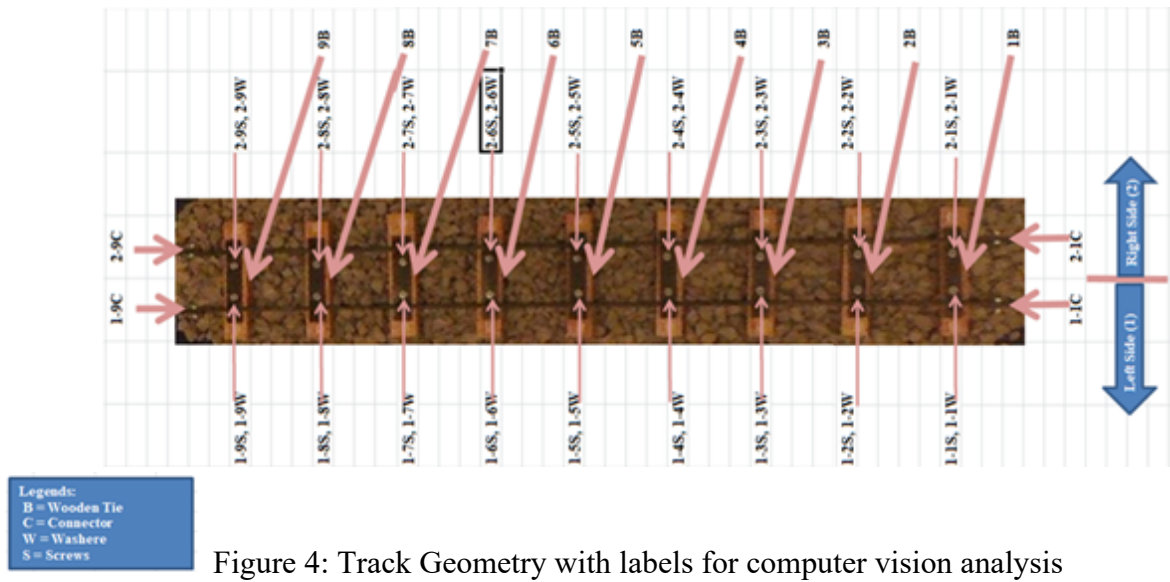

Figure 4: Track Geometry with labels for computer vision analysis

The conducted experimental data was gathered for videos and still images (frames) and tabulated for statistical analysis, results, illustrations and conclusion of this project.

## 4. Preparing Training Datasets for Analysis

The next phase of this experiment was to create the training dataset for state of good tracks (case #01) and analyze the subsequent test cases representing problem tracks using MATLAB Software (case #02 through #15). This was attained by following steps:

4.1) Prepare trained dataset of state of good repairs of rail tracks and their anomalies;

4.2) Acquire images/videos using camera embedded intelligent drone machine and iPhone on simulated track model;

4.3) Pre-processing, using computer vision toolbox on laptop, of acquired images for valid coordinates, noise reduction, enhanced contrast to assure that relevant information is detected, and enhanced image structures gathered for accurate analysis;

4.4) Extract image features at various levels from data collected using lines, edges, ridges and more localized interest points such as corners, blobs or points;

4.5) Detection/segmentation of points of interest with issues/problems such as missing/broken ties, worn out track fastener, missing connectors, etc.;

4.6) High level processing by analyzing the image and points of interest with respect to trained data (supervised learning) for verification so that the data satisfy model and assumptions. This would allow image recognition, registration and comparison with trained dataset;

4.7) Prediction: making the final decision on the rail track safety and security such as "Track is Safe" or "Track is not safe" with visual graphical images showing areas of problems based on trained dataset.

Improving Train Track Safety using Drones, Computer Vision and Machine LearningThe above listed procedures from 4.1 through 4.7 were followed for all 15 test cases whereby the problem tracks were compared with respect to their standard tracks. The results were then tabulated with the testing outcome to define the confidence level of the experiment.

The drone footage in the form of videos captured during track inspections were downloaded for image processing using laptop in ".mp4" format. Footage was taken for five trials to ensure the accuracy and precision of the testing results. They were labeled for record keeping and analysis as tabulated in table – 1: Drone video footage chart (5 trials and 15 test cases) and is listed below:

| Case # | Case Description | Track Defect Locations Expected | Video Names | | | | |
|---|---|---|---|---|---|---|---|
| | | | Trial 1 | Trial 2 | Trial 3 | Trial 4 | Trial 5 |
| 01 | Standard Good Track | Track is Safe: No defects were found | 01_V_T1 | 01_V_T2 | 01_V_T3 | 01_V_T4 | 01_V_T5 |
| 02 | 1 Screw, 1 Washer, 1 Block, 1 Connector missing | Track is NOT Safe: 1-9S, 1-9W, 9B, 1-9C missing | 02_V_T1 | 02_V_T2 | 02_V_T3 | 02_V_T4 | 02_V_T5 |
| 03 | 2 Screws, 2 Washers, 2 Blocks, 2 Connectors missing | Track is NOT Safe: 1-5S, 2-5S, 1-5W, 2-5W, 5B, 9B, 1-9C, 2-9C missing | 03_V_T1 | 03_V_T2 | 03_V_T3 | 03_V_T4 | 03_V_T5 |
| 04 | 1 Screw missing | Track is NOT Safe: 1-7S missing | 04_V_T1 | 04_V_T2 | 04_V_T3 | 04_V_T4 | 04_V_T5 |
| 05 | 2 Screws missing | Track is NOT Safe: 1-7S, 2-4S missing | 05_V_T1 | 05_V_T2 | 05_V_T3 | 05_V_T4 | 05_V_T5 |
| 06 | 1 Washer missing | Track is NOT Safe: 1-3W missing | 06_V_T1 | 06_V_T2 | 06_V_T3 | 06_V_T4 | 06_V_T5 |
| 07 | 2 Washers missing | Track is NOT Safe: 1-3W, 2-7W missing | 07_V_T1 | 07_V_T2 | 07_V_T3 | 07_V_T4 | 07_V_T5 |
| 08 | 1 Block missing | Track is NOT Safe: 1B missing | 08_V_T1 | 08_V_T2 | 08_V_T3 | 08_V_T4 | 08_V_T5 |
| 09 | 2 Block missing | Track is NOT Safe: 2B, 6B missing | 09_V_T1 | 09_V_T2 | 09_V_T3 | 09_V_T4 | 09_V_T5 |
| 10 | 1 Connector missing | Track is NOT Safe: 1-1C missing | 10_V_T1 | 10_V_T2 | 10_V_T3 | 10_V_T4 | 10_V_T5 |
| 11 | 2 Connectors missing | Track is NOT Safe: 2-1C, 2-9C missing | 11_V_T1 | 11_V_T2 | 11_V_T3 | 11_V_T4 | 11_V_T5 |
| 12 | 1 Screw, 1 Washer, 1 Connector missing | Track is NOT Safe: 1-3S, 1-3W, 2-1C missing | 12_V_T1 | 12_V_T2 | 12_V_T3 | 12_V_T4 | 12_V_T5 |
| 13 | 2 Screws, 2 Washers missing | Track is NOT Safe: 1-7S, 2-4S, 1-7W, 2-4W missing | 13_V_T1 | 13_V_T2 | 13_V_T3 | 13_V_T4 | 13_V_T5 |
| 14 | 2 Screws, 2 Washers , 1 Block missing | Track is NOT Safe: 1-8S, 2-8S, 1-8W, 2-8W, 8B missing | 14_V_T1 | 14_V_T2 | 14_V_T3 | 14_V_T4 | 14_V_T5 |
| 15 | 2 Screws, 2 Washers , 1 Block, 1 Connector missing | Track is NOT Safe: 1-8S, 2-8S, 1-8W, 2-8W, 8B, 2-1C missing | 15_V_T1 | 15_V_T2 | 15_V_T3 | 15_V_T4 | 15_V_T5 |

Table 1: Drone Video Footage Trials and their Footage File Names - used for track inspection analysis

Filename example for video: 01_V_T1 – "01"st case, "V"ideo and "T"rail "1". During the same drone trial, still frames/images were taken and were downloaded for image processing using laptop in ".jpg" format. They were labeled for record keeping and analysis as tabulated in "Table 2: Drone frame footage chart (5 trials and 15 test cases)" and is listed next: Filename example for Frames: 01_F_T2 – "01"st case, "F"rame and "T"rial "2"; 01_F_T2 – "01"st case.

| Case # | Case Description | Track Defect Locations Expected | Frame Names | | | | |
|---|---|---|---|---|---|---|---|
| | | | Trial 1 | Trial 2 | Trial 3 | Trial 4 | Trial 5 |
| 01 | Standard Good Track | Track is Safe: No defects were found | 01_F_T1 | 01_F_T2 | 01_F_T3 | 01_F_T4 | 01_F_T5 |
| 02 | 1 Screw, 1 Washer, 1 Block, 1 Connector missing | Track is NOT Safe: 1-9S, 1-9W, 9B, 1-9C missing | 02_F_T1 | 02_F_T2 | 02_F_T3 | 02_F_T4 | 02_F_T5 |
| 03 | 2 Screws, 2 Washers, 2 Blocks, 2 Connectors missing | Track is NOT Safe: 1-5S, 2-5S, 1-5W, 2-5W, 5B, 9B, 1-9C, 2-9C missing | 03_F_T1 | 03_F_T2 | 03_F_T3 | 03_F_T4 | 03_F_T5 |
| 04 | 1 Screw missing | Track is NOT Safe: 1-7S missing | 04_F_T1 | 04_F_T2 | 04_F_T3 | 04_F_T4 | 04_F_T5 |
| 05 | 2 Screws missing | Track is NOT Safe: 1-7S, 2-4S missing | 05_F_T1 | 05_F_T2 | 05_F_T3 | 05_F_T4 | 05_F_T5 |
| 06 | 1 Washer missing | Track is NOT Safe: 1-3W missing | 06_F_T1 | 06_F_T2 | 06_F_T3 | 06_F_T4 | 06_F_T5 |
| 07 | 2 Washers missing | Track is NOT Safe: 1-3W, 2-7W missing | 07_F_T1 | 07_F_T2 | 07_F_T3 | 07_F_T4 | 07_F_T5 |
| 08 | 1 Block missing | Track is NOT Safe: 1B missing | 08_F_T1 | 08_F_T2 | 08_F_T3 | 08_F_T4 | 08_F_T5 |
| 09 | 2 Block missing | Track is NOT Safe: 2B, 6B missing | 09_F_T1 | 09_F_T2 | 09_F_T3 | 09_F_T4 | 09_F_T5 |
| 10 | 1 Connector missing | Track is NOT Safe: 1-1C missing | 10_F_T1 | 10_F_T2 | 10_F_T3 | 10_F_T4 | 10_F_T5 |
| 11 | 2 Connectors missing | Track is NOT Safe: 2-1C, 2-9C missing | 11_F_T1 | 11_F_T2 | 11_F_T3 | 11_F_T4 | 11_F_T5 |
| 12 | 1 Screw, 1 Washer, 1 Connector missing | Track is NOT Safe: 1-3S, 1-3W, 2-1C missing | 12_F_T1 | 12_F_T2 | 12_F_T3 | 12_F_T4 | 12_F_T5 |
| 13 | 2 Screws, 2 Washers missing | Track is NOT Safe: 1-7S, 2-4S, 1-7W, 2-4W missing | 13_F_T1 | 13_F_T2 | 13_F_T3 | 13_F_T4 | 13_F_T5 |
| 14 | 2 Screws, 2 Washers , 1 Block missing | Track is NOT Safe: 1-8S, 2-8S, 1-8W, 2-8W, 8B missing | 14_F_T1 | 14_F_T2 | 14_F_T3 | 14_F_T4 | 14_F_T5 |
| 15 | 2 Screws, 2 Washers , 1 Block, 1 Connector missing | Track is NOT Safe: 1-8S, 2-8S, 1-8W, 2-8W, 8B, 2-1C missing | 15_F_T1 | 15_F_T2 | 15_F_T3 | 15_F_T4 | 15_F_T5 |

Table 2: Drone Frame Footage Trials and their Footage File Names - used for track inspection analysis

The MATLAB Software for image processing and computer vision was used to acquire these images, preprocessed and modeled for final results. Figure 5 shows the experimental design process flow.

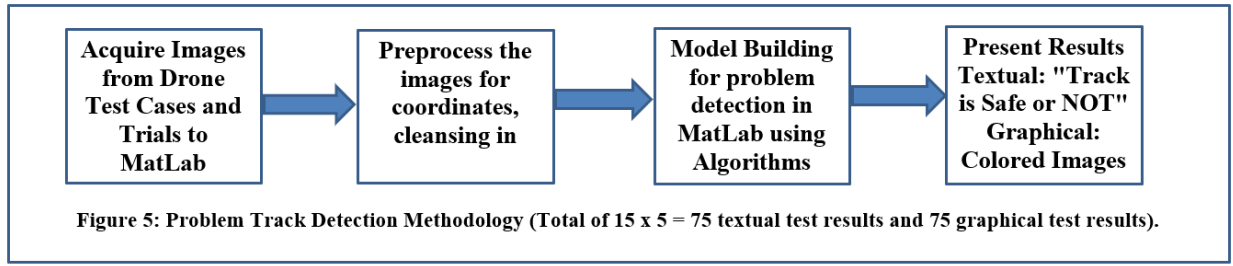

Figure 5: Problem Track Detection Methodology (Total of 15 x 5 = 75 textual test results and 75 graphical test results).

This project was further extended to machine learning approach to extend our curiosity in applying machine learning approaches to identify the track defects.

## Results from Computer Vision Based Model

The images and footage acquired by the drones were processed through the MATLAB computer vision tool program developed for this project on all the test cases. The experimentation results sample is illustrated for case 15 compared to standard track (case 01) and Trial 5: when a standard track footage 01_F_T5.jpg was compared to a problem track footage 15_F_T5.jpg using MATLAB Software, the textual results showed that the track was not safe as shown in Figure 6:

```
Acquiring Image 1 (Control): 01_F_T5.jpg
Acquiring Image 2 (Variable): 15_F_T5.jpg
Pre-processing Image 1 (Control): 01_F_T5.jpg
Pre-processing Image 2 (Variable): 15_F_T5.jpg
Extract Image Features (Control vs. Variable): 01_F_T5.jpg vs. 15_F_T5.jpg
Detection/segmentation of POI (Control vs. Variable): 01_F_T5.jpg vs. 15_F_T5.jpg
Presenting Visual Track Problems (Control vs. Variable): 01_F_T5.jpg vs. 15_F_T5.jpg
   >>Predection of Final Decision: DANGER: ***TRACK IS NOT SAFE!***
```

Figure 6: Textual Results shown when the track is not safe.

The same test result in graphical representation is shown below in Figure 7 where the final picture shows that the track is not safe and has the following problems: 1-8S, 2-8S, 1-8W, 2-8W, 8B and 2-1C missing (S for Screws, W for Washer, C for Connector and B for Block – see methods section for more details). The 1$^{st}$ frame is standard track and the 2$^{nd}$ frame is problem track. After the MATLAB program modeling, the gray scale analysis shows the problem area which was finally presented in RGB scale for track quality.

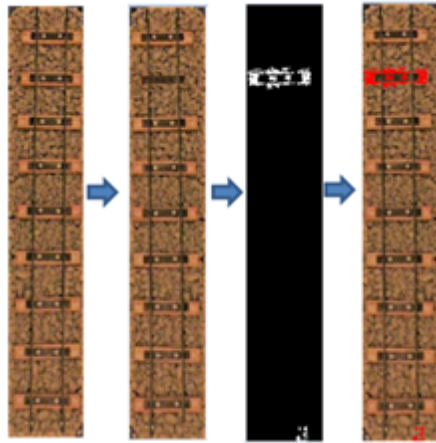

Figure 7: Graphical Test Result displays actual problem spots

The 75 experimental runs (15 Test cases times 5 Trials) were conducted in MATLAB using the following combinations as shown in Table-3 – for each test case and trial, all combination testing were conducted to show the tracks are identified as "Safe" or "Not Safe" with graphical representation of the problem location.

Figure 8 below attempts to illustrate some of the test results in graphical form:

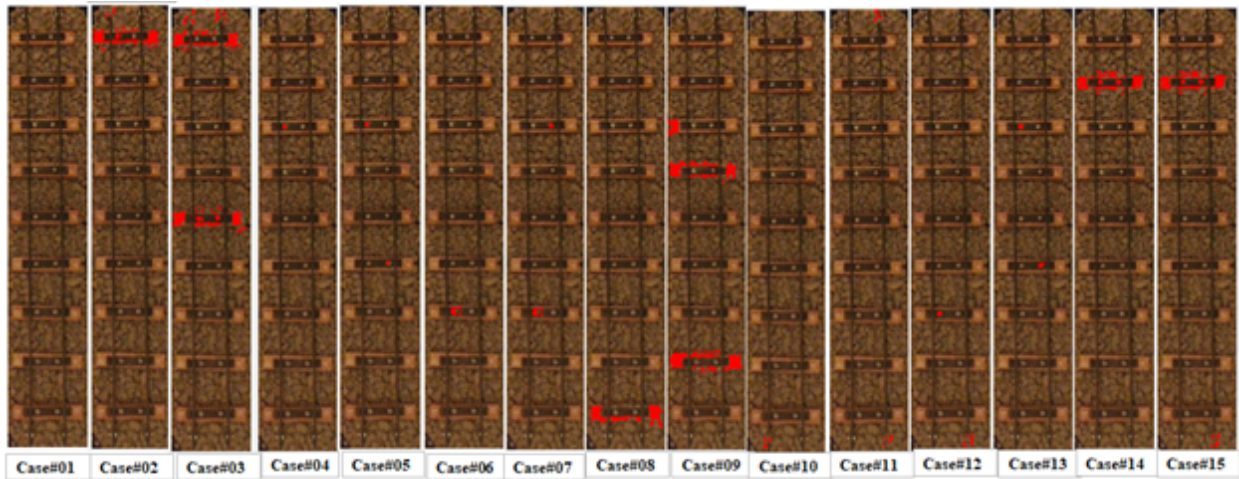

Figure 8: Graphical results of Trial 1 and all test cases show visual track problems

Figure 9 attempts to illustrate a test result in textual form when the track is "Safe":

```
Acquiring Image 1 (Control): 01_F_T1.jpg
Acquiring Image 2 (Variable): 01_F_T5.jpg
Pre-processing Image 1 (Control): 01_F_T1.jpg
Pre-processing Image 2 (Variable): 01_F_T5.jpg
Extract Image Features (Control vs. Variable): 01_F_T1.jpg vs. 01_F_T5.jpg
Detection/segmentation of POI (Control vs. Variable): 01_F_T1.jpg vs. 01_F_T5.jpg
Presenting Visual Track Problems (Control vs. Variable): 01_F_T1.jpg vs. 01_F_T5.jpg
   >>Predection of Final Decision: TRACK IS SAFE
```

Figure 9: Track is safe textual result when standard tracks footage taken in Trial 1 is compared with Trial 5

The variance analysis of track defects observed vs. expected values is shown for each trial and the test case in Table 4. The variance analysis suggests an overall standard deviation of 0.311 and a variance of 0.193. The standard deviation and variance of the data suggest that the data between each trial are spread out showing that there is an improvement needed in the training datasets as well as fault detection algorithms to cause less variation between each trial and more consistent results.

| Case # | Case Description | Track Defect Locations Expected | Trial 1 | Trial 2 | Trial 3 | Trial 4 | Trial 5 |
|---|---|---|---|---|---|---|---|
| 01 | Standard Good Track | Track is Safe: No defects were found | 01_F_T1 / 01_F_T1 | 01_F_T2 / 01_F_T2 | 01_F_T3 / 01_F_T3 | 01_F_T4 / 01_F_T4 | 01_F_T5 / 01_F_T5 |
| 02 | 1 Screw, 1 Washer, 1 Block, 1 Connector missing | Track is NOT Safe: 1-9S, 1-9W, 9B, 1-9C missing | 02_F_T1 / 01_F_T1 | 02_F_T2 / 01_F_T2 | 02_F_T3 / 01_F_T3 | 02_F_T4 / 01_F_T4 | 02_F_T5 / 01_F_T5 |
| 03 | 2 Screws, 2 Washers, 2 Blocks, 2 Connectors missing | Track is NOT Safe: 1-5S, 2-5S, 1-5W, 2-5W, 5B, 9B, 1-9C, 2-9C missing | 03_F_T1 / 01_F_T1 | 03_F_T2 / 01_F_T2 | 03_F_T3 / 01_F_T3 | 03_F_T4 / 01_F_T4 | 03_F_T5 / 01_F_T5 |
| 04 | 1 Screw missing | Track is NOT Safe: 1-7S missing | 04_F_T1 / 01_F_T1 | 04_F_T2 / 01_F_T2 | 04_F_T3 / 01_F_T3 | 04_F_T4 / 01_F_T4 | 04_F_T5 / 01_F_T5 |
| 05 | 2 Screws missing | Track is NOT Safe: 1-7S, 2-4S missing | 05_F_T1 / 01_F_T1 | 05_F_T2 / 01_F_T2 | 05_F_T3 / 01_F_T3 | 05_F_T4 / 01_F_T4 | 05_F_T5 / 01_F_T5 |
| 06 | 1 Washer missing | Track is NOT Safe: 1-3W missing | 06_F_T1 / 01_F_T1 | 06_F_T2 / 01_F_T2 | 06_F_T3 / 01_F_T3 | 06_F_T4 / 01_F_T4 | 06_F_T5 / 01_F_T5 |
| 07 | 2 Washers missing | Track is NOT Safe: 1-3W, 2-7W missing | 07_F_T1 / 01_F_T1 | 07_F_T2 / 01_F_T2 | 07_F_T3 / 01_F_T3 | 07_F_T4 / 01_F_T4 | 07_F_T5 / 01_F_T5 |
| 08 | 1 Block missing | Track is NOT Safe: 1B missing | 08_F_T1 / 01_F_T1 | 08_F_T2 / 01_F_T2 | 08_F_T3 / 01_F_T3 | 08_F_T4 / 01_F_T4 | 08_F_T5 / 01_F_T5 |
| 09 | 2 Blocks missing | Track is NOT Safe: 2B, 6B missing | 09_F_T1 / 01_F_T1 | 09_F_T2 / 01_F_T2 | 09_F_T3 / 01_F_T3 | 09_F_T4 / 01_F_T4 | 09_F_T5 / 01_F_T5 |
| 10 | 1 Connector missing | Track is NOT Safe: 1-1C missing | 10_F_T1 / 01_F_T1 | 10_F_T2 / 01_F_T2 | 10_F_T3 / 01_F_T3 | 10_F_T4 / 01_F_T4 | 10_F_T5 / 01_F_T5 |
| 11 | 2 Connectors missing | Track is NOT Safe: 2-1C, 2-9C missing | 11_F_T1 / 01_F_T1 | 11_F_T2 / 01_F_T2 | 11_F_T3 / 01_F_T3 | 11_F_T4 / 01_F_T4 | 11_F_T5 / 01_F_T5 |
| 12 | 1 Screw, 1 Washer, 1 Connector missing | Track is NOT Safe: 1-3S, 1-3W, 2-1C missing | 12_F_T1 / 01_F_T1 | 12_F_T2 / 01_F_T2 | 12_F_T3 / 01_F_T3 | 12_F_T4 / 01_F_T4 | 12_F_T5 / 01_F_T5 |
| 13 | 2 Screws, 2 Washers missing | Track is NOT Safe: 1-7S, 2-4S, 1-7W, 2-4W missing | 13_F_T1 / 01_F_T1 | 13_F_T2 / 01_F_T2 | 13_F_T3 / 01_F_T3 | 13_F_T4 / 01_F_T4 | 13_F_T5 / 01_F_T5 |
| 14 | 2 Screws, 2 Washers, 1 Block missing | Track is NOT Safe: 1-8S, 2-8S, 1-8W, 2-8W, 8B missing | 14_F_T1 / 01_F_T1 | 14_F_T2 / 01_F_T2 | 14_F_T3 / 01_F_T3 | 14_F_T4 / 01_F_T4 | 14_F_T5 / 01_F_T5 |
| 15 | 2 Screws, 2 Washers, 1 Block, 1 Connector missing | Track is NOT Safe: 1-8S, 2-8S, 1-8W, 2-8W, 8B, 2-1C missing | 15_F_T1 / 01_F_T1 | 15_F_T2 / 01_F_T2 | 15_F_T3 / 01_F_T3 | 15_F_T4 / 01_F_T4 | 15_F_T5 / 01_F_T5 |

Table 3: Drone Frame Footage Analysis - MatLab experimental analysis performed

The test results, both textual and graphical, were manually reviewed to assess the testing results on 1-5 Likert scale according to parameters given by Mogey (1999) and Table 5 shows the overall average success rate of 96.09% in identifying track defects. This average value was computed for each test case for all the trials and then averaged for all test cases' average value and the formula is shown below (the test results' variability is between 4 and 5 and hence average options was selected over Kruskal Wallis test (Mogey, 1999):

$$\text{Overall Likert Scale Acceptance Average} = \frac{\sum_{\text{Test Cases } C=1}^{C=15} \left( \sum_{\text{Trial } T=1}^{T=5} \frac{\text{Likert Scale Value (T)}}{\text{Total Trials}=5} \right)}{\text{Total Test Cases } = 15 \ast \text{Likert scale}=5}$$

Figure 10: Likert scale formula

The frequency distribution of Likert scale suggests that the Likert scale rating occurrence range between 4 and 5 with maximum frequency of 5 dominates indicating higher success rate identifying track defects, supported by the frequency regression lines' downward trend and is shown in Figure 11:

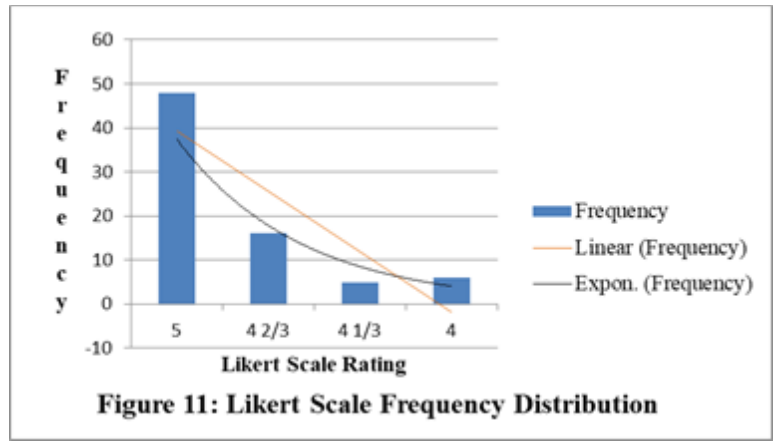

Figure 11: Likert Scale Frequency Distribution

The rubrics for the Likert scale assessment is listed below: (i) If the textual result showed properly "Safe" or "Not Safe", then a score of 1 will be given out of 5; (ii) If the textual result showed all the steps executed properly, then an additional score of 1 will be given out of 5; (iii) If the graphical result showed all false negative conditions, then an additional score of 1. If the false negative is partial $1/3^{rd}$ if shown low, $2/3^{rd}$ if shown medium and 1 if shown high (all of the track problems); (iv) If the graphical result showed all the steps executed properly, then an additional score of 1 will be given out of 5; (v) If textual and graphical results concur, then an additional score of 1 will be given out of 5; (vi) If the graphical result showed all false positive conditions, then an additional score of -1 will be given out of 5. If the false positive is partial $-1/3^{rd}$ if shown low, $-2/3^{rd}$ if shown medium and -1 if shown high (problems are shown predominantly where not expected).

Please note that Likert scale rubrics can be improvised by conducting a survey with safety and security engineers currently performing track inspections. By applying the above rubrics for the Likert scale to this project, the test results showed an overall average of 96.089% acceptable rate in identifying track defects. However, this acceptable rate can be further improved by refining the MATLAB algorithms to improve track problems identification and reduce false positives.

The Pseudocode for the MATLAB algorithm for identifying the track defects is documented below:
  i. Initialize the images to compare (source and target);
  ii. Create the filename with images compared for the textual results storage and open.
  iii. Create the filename with images compared for the graphical results storage and open.
  iv. Acquire the images to compare (source and target) and register in the test results.

v. Pre-process the images from RGB to BW scales for cleansing and further analysis; register the test results.
vi. Extract image features for identifying the problem areas; register the test results.
vii. Analyze the image by comparing special features for detecting problem areas; register the test results.
viii. Present the visual graphical display of target track compared with respect to the source track with possible issues/problems. Display in RGB scale with red colors indicating problem areas.
ix. Finally, provide the textual results of "TRACK IS SAFE" or "DANGER: ***TRACK IS NOT SAFE!***" depending on the track inspection results. Then, close the program so that testing results are stored for further analysis.

The graphical representation of Standard Deviation of the test case results of identifying the track defects is shown in Figure 12. The graph suggests that in many test cases the results were acceptable but in several test cases the standard deviation was high leaving a standard deviation of 0.311 for the experiment as a whole. This indicates further refinement is necessary in the experimental design to have better standard deviation and less variance.

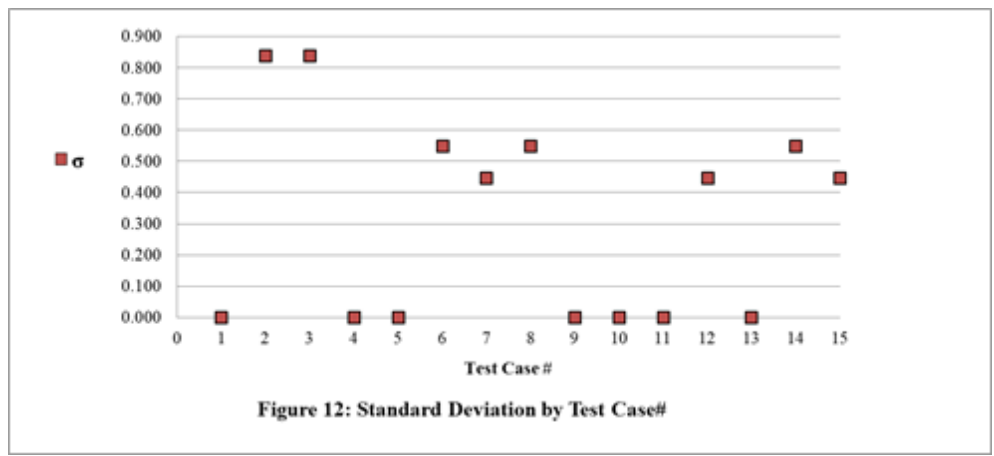

Figure 12: Standard Deviation by Test Case#

The graphical representation of standard deviation of the test case results of identifying the track defects (figure 12) suggests test results have a high standard deviation and have a high acceptable rate and deviation.

The false positives using Likert scale (figure 13) also suggests high acceptable rate of 5 and hence the approach is feasible.

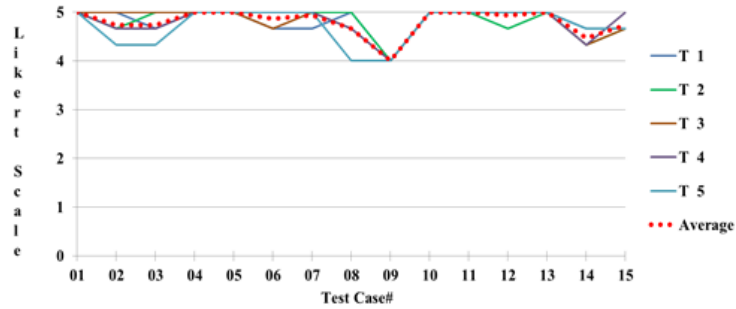

Figure 13. False Positive by Track Estimated Using Likert Scale

The confusion matrix in Table 4 demonstrates a high acceptable TPR of 98.84% indicating the success of this project. However, the FPR value of 7.88% indicates some scope for improvement. The figure 14 demonstrates the ROC analysis for threshold determination indicating a very high acceptance rate with threshold of 10 during the computer vision analysis.

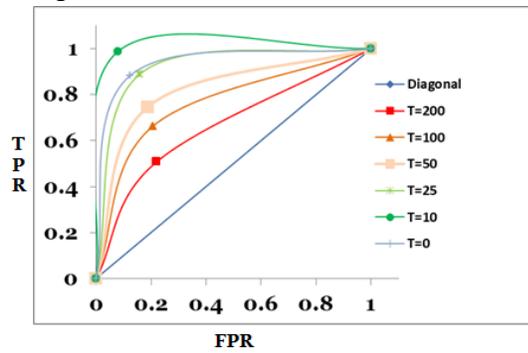

Figure 14. Receiver Operating Characteristics (ROC) Curve for threshold determination

TP = True Positive
FN = False Negative
FP = False Positive
TN = True Negative

| Confusion Matrix | Defect Reported | Defect Not Reported |
|---|---|---|
| Defect Present | 171 TP | 2 FN |
| Defect Absent | 23 FP | 269 TN |

True Positive Rate (TPR) = $\frac{TP}{(TP + FN)} = \frac{171}{171 + 2} = 98.84\%$

False Positive Rate (FPR) = $\frac{FP}{(FP + TN)} = \frac{23}{23+269} = 7.88\%$

Table 4. Confusion Matrix Analysis

# Design and Implementation using Machine Learning

The same test cases were further verified using the AI/ML techniques by using the four-step process: 1) Obtain and prepare the images from Drones (same footage and images taken from Drone with the MATLAB/Computer Vision approach; 2) Build and compile the machine learning model; 3) Train and evaluate the model; and finally 4) Test the model and summarized the results.

The footage acquired from drones were classified into good and bad tracks based on the test cases with supplemented images using image augmentation techniques to have sizable sample datasets to allow the models to function (used Keras Image Data Generators). The images were classified into training datasets with good and bad tracks and the results were tracked in the validation and testing folders (about 400 images in the training folder, 200 into the validation folder and 200 into the testing folder).

Next, a model was acquired for VGG16, an existing model used to classify images into a thousand objects that won the ImageNet Large Scale Visual Recognition Challenge in 2014 (to identify cats/dogs). This model was adapted to detect track defects based on the images that were classified earlier in the first step, the concept called transfer learning. These steps are shown in Figure 5A.

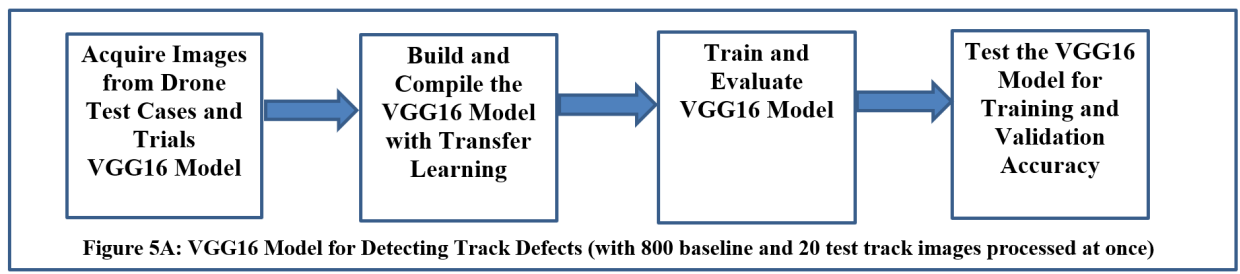

Figure 5A: VGG16 Model for Detecting Track Defects (with 800 baseline and 20 test track images processed at once)

In addition, this project built a custom model using the Keras neural network API in python along with a TensorFlow backend to support track defect detection. Keras is an open-source neural network library and TensorFlow is Google's open source library for large-scale machine learning and computation. For this custom model, we used three collated 'blocks' of Convolution, Activation, and Pooling layers, followed by Flatten and Dense layers to reduce the dimension of the output to 2 options (safe and defective). Initial layers were used to help the model detect smaller shapes, such as lines or curves, while the later layers to identify more complex structures, with the final layer making the ultimate decision. The compilation was performed using the optimizer, which determines whether the model is learning (Adam). The loss function, which was used to calculate the extent of the gap between predicted and actual classifications (Categorical Cross Entropy), and the metrics for training accuracy (decimal representation of the fraction of training set predictions that were correct) and validation accuracy (decimal representation of the fraction of validation set predictions that were correct). These steps are shown in Figure 5B.

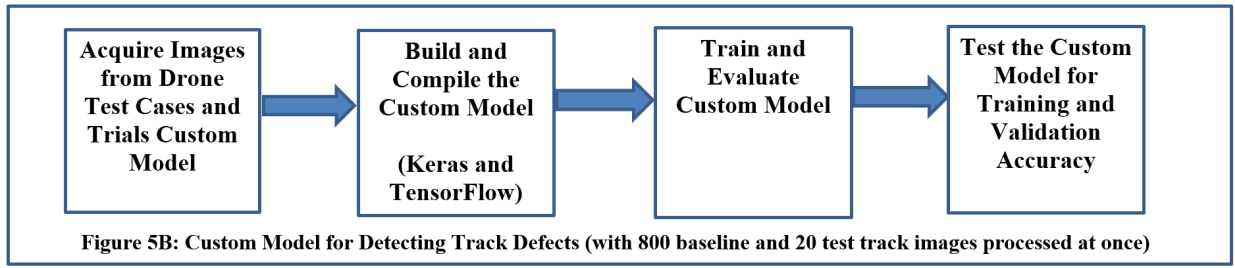

Figure 5B: Custom Model for Detecting Track Defects (with 800 baseline and 20 test track images processed at once)

The next step was to train and evaluate these two models - Custom Model and VGG16 (transfer learning). This project created an iterator for each of the three folders of images (train, valid, and test), using Keras Image Data Generators, to feed the images in those folders into the model. These iterators can specify several parameters, such as batch size (how many pictures enter the model at a time), target image size, and shuffle (randomizes the images pulled from the folder). A large batch size is quicker, because the model has to be updated less frequently, but it requires more memory to handle many images at once. This project used a central number of 20 as our batch size to avoid either extreme and manage performance (a very tedious task when performance is an issue). Training of the models were done using the fit generator function on the Keras module. The values for the number of epochs (the number of times the data was run through the model to train it), steps per epoch (the number of batches fed into the model), and validation steps (same as steps per epoch, but for the validation data) were manually provided. The evaluation of the important output metrics, namely loss and accuracy for both the training and validation data sets, were managed by utilizing the history object to track the progress of the model. Plotting this object helped in visual inspection of the accuracy and loss. The validation images were used to test the model after each epoch. Thus, the training accuracies and validation accuracies were the best metrics to demonstrate the model validity. The optional model was attained by changing the number of layers, batch size, and optimizer along with other parameters.

The final step was to test these models and summarize our results. The test generator used an iterator to feed the test images (real-time images taken from the drones for testing) into the model to predict whether inspected track images were safe or defective. The results were captured and plotted on a confusion matrix to display the number of true positives, false positives, true negatives, and false negatives. These confusion matrix results were plotted for accuracy and epochs for training and validation results.

Initially, poor results were received for both models when using all types of defects (missing blocks, screws/washers and connectors). Testing was performed to assess if the program was categorizing only one type of defect correctly and not for the others, so we created multiple versions of the models to detect specific problem such as wooden blocks missing, screws/washers missing, and/or connectors missing. The models were well trained after conducting thousands of iterations and combinations to various problem types by tweaking various options provided in the custom model and VGG16 model.

## Results from Transfer Learning Model

The testing results for the VGG16 model on defective tracks with all types of defects provided a confusion matrix which indicated a moderate false negative (23%) and high false positive (36%) in identifying the track defects. This model provided a 62% training accuracy and 69% validation accuracy. In addition, the validation accuracy wildly fluctuated, returning highly inconsistent results, demonstrating the unsuitability of this approach. The results are shown in Figure 15.

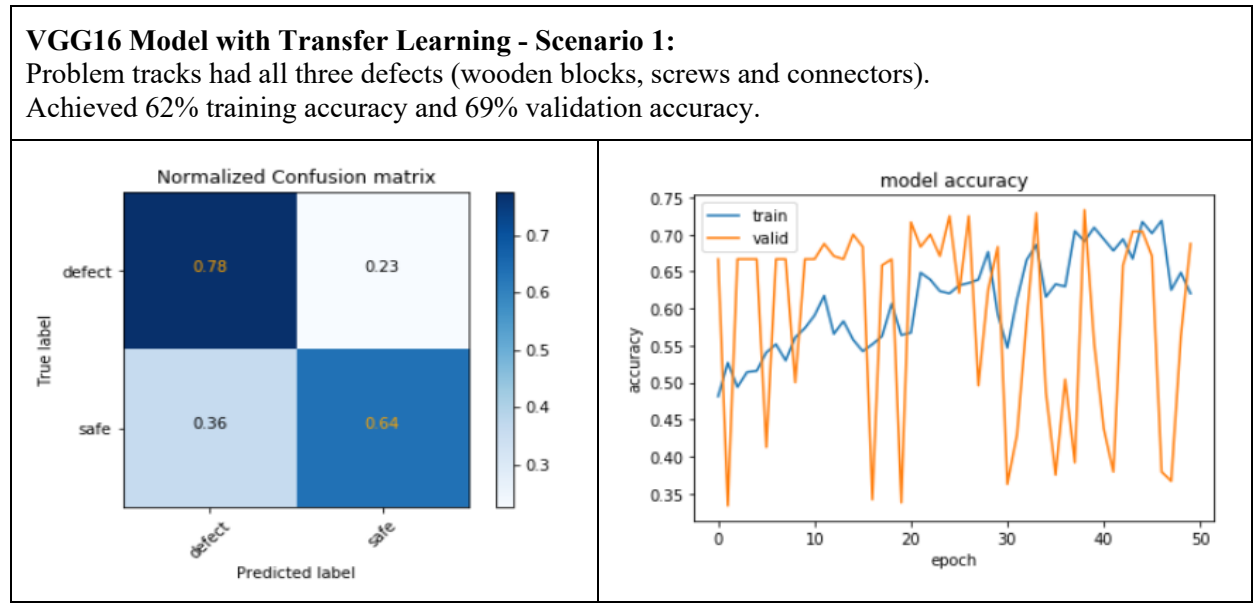

Figure 15: Scenario 1 - VGG Model Results (Confusion Matrix and Model Accuracy)

This resulted in conducting additional scenarios to further analyze if the VGG16 model would be viable if only one defect were to be introduced during our testing (missing wooden blocks). The results are shown below suggests that the model achieved 97% training accuracy and validation accuracy of 73%. As the confusion matrix indicates the number of false negatives was small while the number of false positives was negligible. The poor validation accuracy combined with the high training accuracy suggest a problem of overfitting, when the model does not learn to identify key features and instead memorizes the input dataset to a degree. These results are shown in Figure 16.

**VGG16 Model with Transfer Learning - Scenario 2:**
Problem tracks had wooden block defects only.
Achieved 97% training accuracy and validation accuracy of 73%.

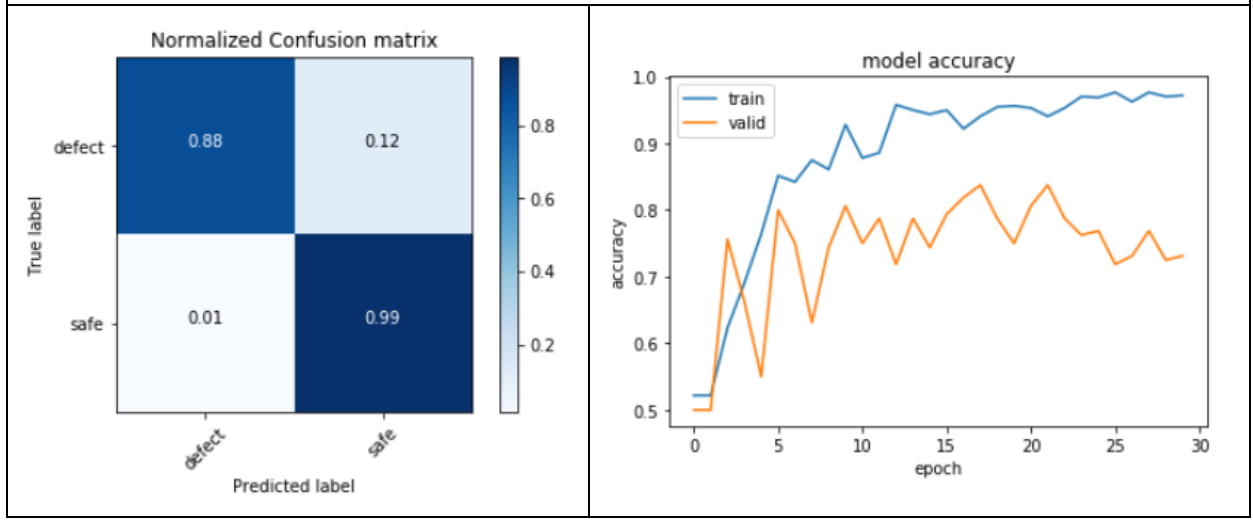

**Figure 16: Scenario 2 - VGG Model Results (Confusion Matrix and Model Accuracy)**

Since the VGG16 model did not provide a convincing machine learning solution using the transfer learning along with a time-intensive execution performance (2-3 hours rather than minutes like in MATLAB), this project aimed to build a custom model to detect track defects using machine learning (with Keras and TensorFlow).

## Results from Custom Machine Learning Model

The testing results with the custom model on defective tracks with all types of defects provided a confusion matrix indicating a high false negative (42%) and high false positive (44%) in identifying the track defects. This model provided a very high 99% training accuracy and a very low 57% validation accuracy (which the project considered unacceptable). This model showed an even greater case of overfitting, in which the model was only able to memorize the input data (resulting in an extremely high training accuracy) but was not able to actually learn anything (resulting in an extremely low validation accuracy). The results are shown in Figure 17.

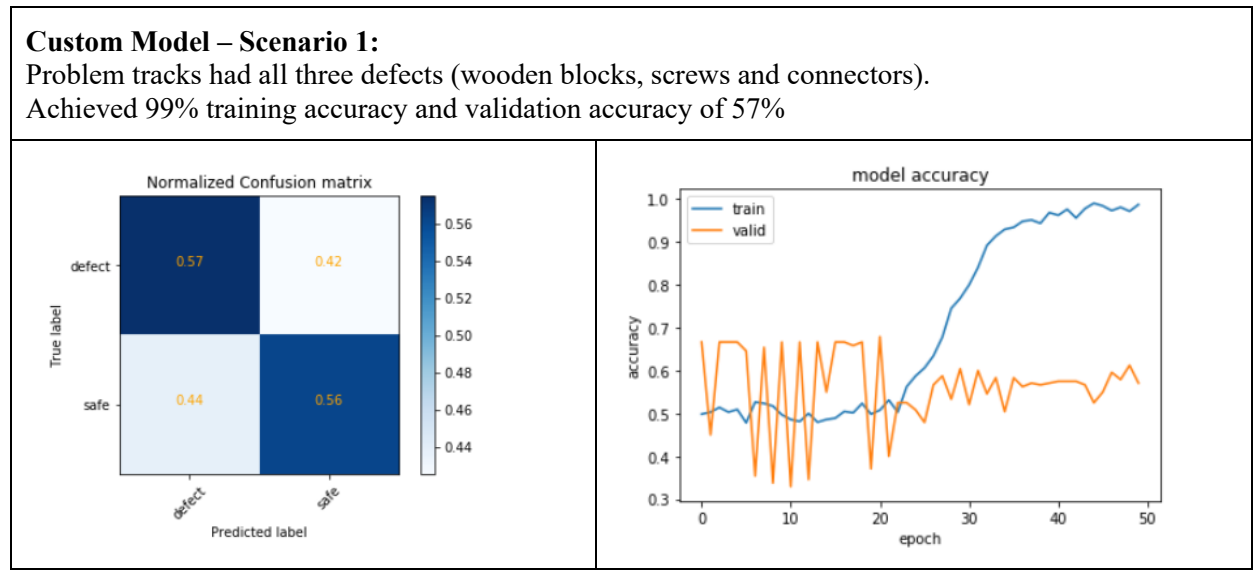

**Custom Model – Scenario 1:**
Problem tracks had all three defects (wooden blocks, screws and connectors).
Achieved 99% training accuracy and validation accuracy of 57%

**Figure 17: Scenario 1 - Custom Model Results (Confusion Matrix and Model Accuracy)**

This resulted in conducting an additional scenario to further analyze if the custom model would be viable if only one defect was introduced during our testing (missing wooden block). The results show that the model achieved 94% training accuracy and validation accuracy of 85%. As the confusion matrix indicates the number of false negatives was low (14%) with slightly fewer false positives (12%). The test results are shown in Figure 18.

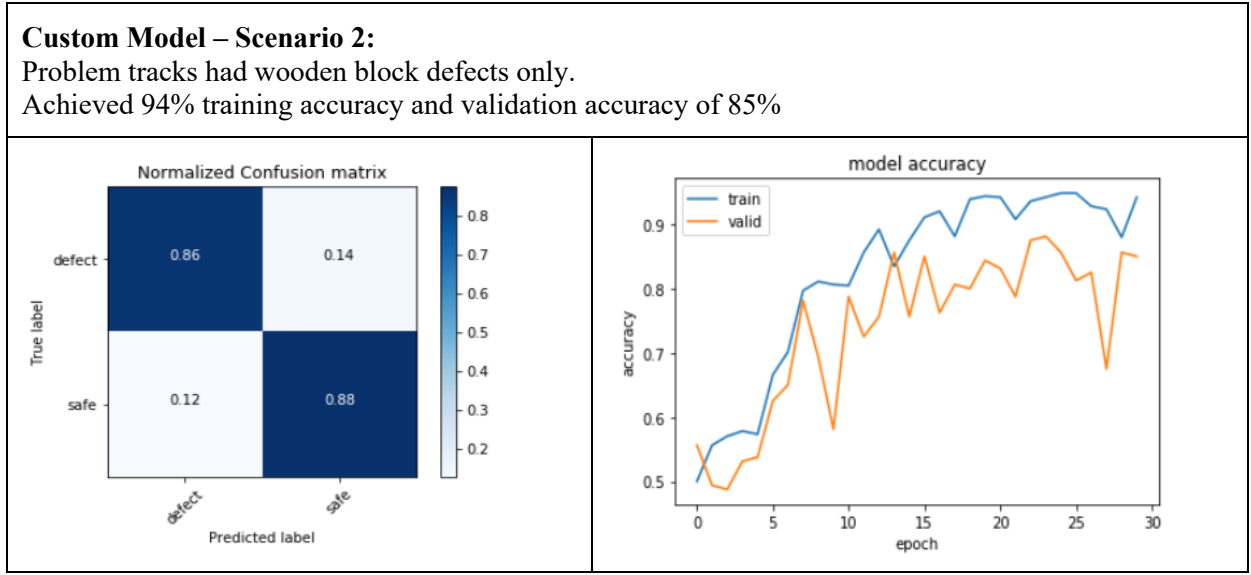

**Custom Model – Scenario 2:**
Problem tracks had wooden block defects only.
Achieved 94% training accuracy and validation accuracy of 85%

Figure 18: Scenario 2 - Custom Model Results (Confusion Matrix and Model Accuracy)

We conducted an additional scenario to further analyze if the custom model is still viable if a different type of defect was used during our testing instead (missing screws). The results show that the model achieved a 98% training accuracy and a validation accuracy of 89%. As the confusion matrix indicates, the false negatives were low at 10% with a minimal number of false positives. The test results are shown in Figure 19.

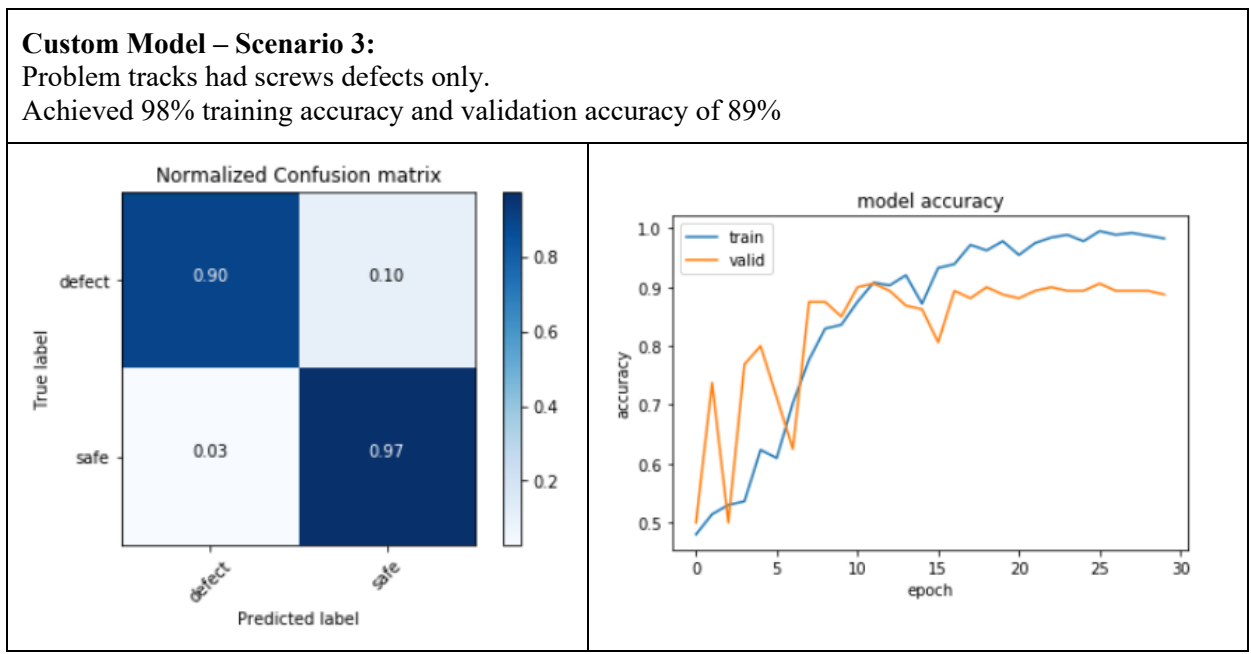

**Custom Model – Scenario 3:**
Problem tracks had screws defects only.
Achieved 98% training accuracy and validation accuracy of 89%

Figure 19: Scenario 3 - Custom Model Results (Confusion Matrix and Model Accuracy)

The results from the custom model demonstrated a promising future for identifying track defects using machine learning with acceptable and expedient performance of testing results.

# Conclusion

The safety and security of railroad transportation industry is under tremendous pressure because of the nonstop occurrences of accidents resulting in derailments due to track issues. The inspection of these track systems is conducted manually by visual inspections once every other week or a month. It is extremely difficult to perform any trend analysis when the inspections are manual and subjective in nature across the vast majority of these train tracks and their continuous expansions. It is very difficult to maintain records on these manual inspections and proactively maintain a state of good repair on track structures. At the same time, these manual inspection results are highly subjective in the interpretation of track structures which varies from one inspector to another. This project demonstrates the ability to apply the use of drones and computer vision techniques to constantly monitor the train track structure and apply artificial intelligence algorithms so that track safety and security is assured cost effectively on daily basis rather than once a month or once in two weeks. This project also demonstrated a consistent way to maintain the inspection record keeping objectively that can be further analyzed for trend analysis and eliminate subjective element in assessing the rail track state of good repairs. This will help the safety and security engineers to be more proactive in problem solving rather than the current status of reactive care.

The project was started with using computer vision (MATLAB) to identify the track defects resulting in the 96.09% success rate, both graphical and textual results in identifying the problem along with True Positive Rate (TPR) of 98.84% indicating the success of this project. However, the False Positive Rate (FPR) value of 7.88% indicated some scope for improvement. This experimentation had several challenges and required future work to overcome these challenges. This experimentation required consistency of drone's flight path and thus resulted in thousands of trials to get a few usable videos. This resulted in our extension to identify track defects using machine learning, implementing both VGG16 and a Custom model (using Keras and TensorFlow) for identifying track defects, adopting to multi-layered Convolution Neural Networks (CNNs). The results from VGG16 achieved 69 to 97% training accuracy and 69 to 73% validation accuracy with long execution times (ran for 2-3 hours for the sample dataset processed). The high training accuracy coupled with the low validation accuracy suggested a dire case of overfitting, in which the model simply memorized the input images instead of actually learning the differences between the safe and defective track. However, the Custom model demonstrated an incredible promise to identify individual track defects, achieving 94 to 98% training accuracy and 85 to 89% validation accuracy with acceptable execution times (within minutes).

Finally, this project demonstrated that with the use of drones (intelligent machines), computer vision and machine learning algorithms (artificial intelligence) with training datasets, the track defects can be identified objectively for maintaining safety and security of rail track maintenance with visual and textual aid support (supervised learning). These algorithms can be further improved to ensure highest level of safety and security in the transportation industry that could

# Improving Train Track Safety using Drones, Computer Vision and Machine Learning

prevent millions of casualties and injuries, improve commuter satisfaction and their confidence in public transportation, increase revenue, protect public/private investments, and ensure a cleaner environment. The project team is executing the hybrid approach of computer vision and the Custom model to demonstrate visual (graphical outputs) and contextual track defect identification can be attained in real-time within reasonable execution times to revolutionize the track inspection marketplace to be on an autonomous state of good repairs.

This project is critical to our local community, the DC Metro area, as well as most cities depending on the metro transportation. The global impact is apparent with the various train accidents that occur globally, impacting millions of lives and causing losses of billions of dollars due to massive destruction. This project demonstrated that the state of good repairs in railway tracks can be attained through computer vision algorithms (supervised learning) that can support the continuous monitoring and analyzing of computer images captured by drones with trained datasets. Through further research, this project can be extended to semi-supervised and unsupervised machine learning. This project will increase public confidence in railways and local rail agencies, counteract the increases in pollution that plague most of the metropolitan cities today, and reduce climate change. The multiple applications and community impact make this project crucial in the real world. Therefore, many improvements can be made to further our study. Additional technologies in drones such as infrared, ultrasound, and sonar capabilities must be enabled so that comprehensive track inspection can be performed for all parts of the track systems such as track gauge, track structure wellness, tie composition stability, power systems supporting rail, switches, wedges, etc. including the underlying track ground density. This project needs further research related to processing time and storage due to the nature of the computer vision and artificial intelligence application. The image processing requires significant computing horsepower and large amount of storage because of video and frames (images) file sizes being large. Daily inspection may need large amount of space to store and high-end computing to perform analysis on inspection data.

There are several factors that must be accounted for the effectiveness of this project. The drone's access to Wi-Fi technology may be a potential issue in the underground tunnel systems and may require some technology upgrades across the current railroad infrastructure. Debris and lighting conditions may distort the inspection outcome and hence require further considerations. The testing was performed under standard house lighting condition in the basement that may require further consideration based on different lighting conditions such as tunnel lighting, sunlight and others. This can always be supplemented with special lighting in drones to create standard lighting conditions. Additionally, implementation of augmented virtual reality (AVR) and embedded sensor systems in track structures to collect multiple datasets supporting the monitoring and analyzing the track systems. We can also further extend these algorithms to semi-supervised and unsupervised machine learning to automate all parts of track inspections and create autonomous monitoring of track safety. This will increase monitoring efficiency and reduce the expenditures

**Improving Train Track Safety using Drones, Computer Vision and Machine Learning**

of rail/metro organizations with better record keeping that can be analyzed and assessed for supervised inspections by safety engineers. This project has the potential for other industries such as the airline, automotive, maritime transport, and autonomous vehicles to redefine safety and security within a predefined environment/sites promoting environment-friendly, safer and secure smart cities.

**Improving Train Track Safety using Drones, Computer Vision and Machine Learning**

# Acknowledgements:

We would like to thank our robotics and automation teacher Mr. Charles DelaCuesta for his mentorship and approval for this project. He provided support in providing the lab access to conduct the experiments and print 3D models of the train tracks for prototype development. We would like to thank Washington Metropolitan Area Transit Authority (WMATA) for providing their research and development opportunity to study the track defects (especially Mr. Andy Off, Chief of Rails and Ms. Laura Mason, Senior vice president of Rail Services at WMATA), reviewing their safety and security procedures, maintenance review of track inspections and many other principles that covered the train track structures. Lastly, we would like to thank our families for their continuous support, patience throughout this project and countless damages done to our basement, drones and simulated track structures during our experiments.

## Bibliography, References and Acknowledgements

- Alpaydin, E. (2016). Machine Learning. Cambridge, Mass.: MIT Press.
- Association of American Railroads. (n.d.). State of The Industry Reports: Report 1 - Safety and Innovation (E. R. Hamberger, Author).
- Babenko, P. (2006). Visual Inspection of Railroad Tracks.
- Belkhade, A., & Kathale, S. (n.d.). AUTOMATIC VISION BASED INSPECTION OF RAILWAY TRACK: A REVIEW.
- Bertram, D. (n.d.). Likert Scales. Retrieved from http://poincare.matf.bg.ac.rs/~kristina/topic-dane-likert.pdf
- Bitar, R. (2015). WMATA - 1000 Track Maintenance & Inspection Procedures.
- Brownlee, J. (2016, June 20). Dropout Regularization in Deep Learning Models With Keras. Retrieved December 10, 2018, from https://machinelearningmastery.com/dropout-regularization-deep-learning-models-keras/
- Camargo, L. F. M., Resendiz, E., Hart, J., Edwards, R., Ahuja, N., & Barkan, C. (2011, January). Machine Vision Inspection of Railroad Track.
- Chollet, F. (2016, June 5). Building powerful image classification models using very little data [Blog post]. Retrieved from The Keras Blog: https://blog.keras.io/building-powerful-image-classification-models-using-very-little-data.html
- Chun, A. H., & Suen, T. Y. (n.d.). Engineering Works Scheduling for Hong Kong's Rail Network.
- Edwards, J. R., & Hart, J. M. (2009). Advancements in Railroad Track Inspection.
- Federal Railroad Administration. (2017, January). New Ideas for Rail Safety (IDEA).
- Federal Railroad Administration Office of Safety. (2006, June). Accident/Incident Data through June 2017. Retrieved from http://safetydata.fra.dot.gov/officeofsafety/publicsite/on_the_fly_download.aspx?itemno=3.03.
- Fine-tune VGG16 Image Classifier with Keras | Part 1: Build [Video file]. (2017, November 22). Retrieved from https://www.youtube.com/watch?v=oDHpqu52soI
- Fine-tune VGG16 Image Classifier with Keras | Part 2: Train [Video file]. (2017, November 22). Retrieved from https://www.youtube.com/watch?v=INaX55V1zpY
- Fine-tune VGG16 Image Classifier with Keras | Part 3: Predict [Video file]. (2017, November 22). Retrieved from https://www.youtube.com/watch?v=HDom7mAxCdc
- Freid, B., Barkan, C. P. L., Ahuja, N., Hart, J. M., Tordorvic, S., & Kocher, N. Multispectral Machine Vision for Improved Undercarriage Inspection of Railroad Rolling Stock.
- Gibert, X., Patel, V. M., & Chellappa, R. Deep Multi-task Learning for Railway Track Inspection.
- Ghouzam, Y. (2017, July 18). Introduction to CNN Keras - Acc 0.997 (top 8%). Retrieved December 10, 2018, from https://www.kaggle.com/yassineghouzam/introduction-to-cnn-keras-0-997-top-6/notebook
- Gu, J., Wang, Z., Kuen, J., Ma, L., Shahroudy, A., Shuai, B., . . . Chen, T. (2018). Recent advances in convolutional neural networks. Pattern Recognition, 77, 354-377. https://doi.org/10.1016/j.patcog.2017.10.013
- Hands-On Convolutional Neural Networks with TensorFlow: Solve computer vision problems with modeling in TensorFlow and Python by Iffat Zafar, Giounona Tzanidou Richard Burton, Nimesh Patel and Leonardo Araujo (2018, August) www.packt.com
- Image Processing and Computer Vision - Geospatial Computing. (n.d.). Retrieved from MATLAB website: https://www.mathworks.com/solutions/image-video-processing/geospatial-computing.html
- Kelleher, J. D., Mac Namee, B., & D'Arcy, A. (2015). Fundamentals of machine learning for predictive data analytics: Algorithms, worked examples, and case studies. Cambridge, MA: The MIT Press.
- Liu, X., Saat, R., M., & Baarkan, C. (n.d.). Analysis of Causes of Major Train Derailment and Their Effect on Accident Rates.
- Mogey, N. (1999, March 25). So you want to use a Likert Scale? Retrieved August 10, 2017, from http://www.icbl.hw.ac.uk/ltdi/cookbook/info_likert_scale/index.html

# Improving Train Track Safety using Drones, Computer Vision and Machine Learning

- M. Muja and D. Lowe. U.S. Federal Transit Administration. (2008, October). Transit State of Good Repair: Beginning the Dialogue. "Fast approximate nearest neighbors with automatic algorithm configuration," in International Conference on Computer Vision Theory and Application VISSAPP'09). pp. 331-340.
- Nageshwaran, A. S., & Geetha, T. S. (2014, April). Wheelbot - An Integrated Robotic Rail Track Inspection and Surveillance System.
- National Transportation Safety Board (NTSB). (2016). Annual Report to Congress.
- Saarenketo, T., and T. Scullion. Texas Transportation Institute, the Texas A&M University System, College Station, TX, 1994. Ground Penetrating Radar Applications on Roads and Highways. Research Report 1923-2F.
- Srivastava, N., Hinton, G., Krizhevsky, A., Sutskever, I., & Salakhutdinov, R. (2014). Dropout: A Simple Way to Prevent Neural Networks from Overfitting. Journal of Machine Learning Research, 1929-1958. Retrieved from http://jmlr.org/papers/volume15/srivastava14a/srivastava14a.pdf
- Stanford University School of Engineering. (2017, August 11). Lecture 1 | Introduction to Convolutional Neural Networks for Visual Recognition [Video file]. Retrieved from https://www.youtube.com/watch?v=vT1JzLTH4G4&list=PL3FW7Lu3i5JvHM8ljYj-zLfQRF3EO8sYv
- Szeliski, R. (2010). Computer Vision: Algorithms and Applications.
- Transportation Safety Board (TSB) of Canada. (2014, March). Statistical Summary, Railway Occurrences 2012. Retrieved from http://www.tsb.gc.ca/eng/stats/rail/2012/ss12.asp.
- University of St. Andrews. (n.d.). Analysing Likert Scale/Type Data, Ordinal Logistic Regression Example in R. Retrieved from https://www.st-andrews.ac.uk/media/capod/students/mathssupport/OrdinalexampleR.pdf
- US Department of Transportation Bureau of Transportation Statistics. (2016). Transportation Statistics Annual Report (M. Bronzini, J. Camp, W. Fletcher, C. Rick, & J. Sedor, Authors).
- US Department of Transportation Federal Railroad Administration. (2013, February). Federal Railroad Administration Track Safety Standards Fact Sheet.
- US Department of Transportation Federal Railroad Agency. Embracing Technology for Railroad Track Inspection (G. Carr, Author).
- US Department of Transportation Federal Railroad Agency. Robust Anomaly Detection for Vision-Based Inspection of Railway Components.
- US Department of Transportation Federal Railroad Administration. (1998). Railroad Safety Report.
- US Department of Transportation Federal Railroad Administration. (2010). Railroad Safety Statistics.
- US Department of Transportation Office of the Assistant Secretary of Research and Technology. (2015, March). An Automated System for Rail Transit Infrastructure Inspection.
- Wang, Y., Zhai, J., Li, Y., Chen, K., & Xue, H. (2018). Transfer learning with partial related "instance-feature" knowledge. Neurocomputing, 310, 115-124. https://doi.org/10.1016/j.neucom.2018.05.029